# PRUNNIG ALGORITHM OF GENERATION A MINIMAL SET OF RULE REDUCTS BASED ON ROUGH SET THEORY

Şahin Emrah AMRAHOV
*Computer Engineering Department , Ankara University, Ankara, Turkey*
*Fatih AYBAR*
*Computer Engineering Department, Ankara University, Ankara, Turkey*
*Serhat DOĞAN*
*Economics Department, Bilkent University, Ankara, Turkey*

In this paper it is considered rule reduct generation problem, based on Rough Set Theory. Rule Reduct Generation (RG) and Modified Rule Generation (MRG) algorithms are well-known. Alternative to these algorithms Pruning Algorithm of Generation A Minimal Set of Rule Reducts, or briefly Pruning Rule Generation (PRG) algorithm is developed. PRG algorithm uses tree structured data type. PRG algorithm is compared with RG and MRG algorithms.

## 1. INTRODUCTION

Nowadays as working area and specialization increase, obtained amount of information also increases comparatively. Lately it becomes necessary to interpret information sets and getting results from them. In this topic Rough Sets Theory is used as an important tool for discovering information from large data sets. Rough Sets Theory is developed by Pawlak ( 1982) and it is applied in many areas. Some of these areas are medical diagnosis, (Wakulicz-Deja and Paszek 1997; Slowinski K. et. al. 2002), artificial intelligence (Lingras,1996), finance (Mrozek and Ekabek ,1998), conflict resolution (Pawlak ,1984), image analysis (Mrozek and Plonka ,1993), pattern recognition,(Manila et. al. 1984; Griffin and Chen,1998; Slowinski and Stefanowski ,1989), control theory (Pawlak and Munakata, 1996), feature extraction (Kusiak and Tseng 1999; Kusiak 2000), classification and rule reduction (Grzymala-Busse and Wang ,1996; Khoo et. al. 1999), machine learning ( Ziarko 1993; Yao et. al. ,1997) and expert systems ( Grzymala-Busse 1991; 1992). One of the areas in which Rough Sets Theory is used is classification and rule reduction. The first algorithm Rule Reduct Generation (RG) is proposed by Pawlak (1991) and modified by Kusiak and Tseng (1999). RG algorithm includes important deficiency. The algorithm examines all the situations and it considers all rules that it found as rule reduction. The second algorithm Modified Rule Generation (MRG) is developed by Guo and Chankong (2002) as modified of RG. This algorithm fills the deficiency of RG, but in order to achieve this, information system has to be reorganized before each examination. In this paper, study on Pruning Algorithm of Generation A Minimal Set of Rule Reducts, or briefly Pruning Rule Generation (PRG) algorithm aim of which is finding the minimum number of rule reduct situations is explained. Different from MRG, this algorithm uses tree structured data type. In the first two part of this study, in which rule reduct algorithms used in Rough Sets Theory are explained, RG and MRG algorithms are given. In the third part, PRG algorithm which is developed in this paper alternatively to these two methods is declared. There is a comparison between these algorithms that uses a sample decision table in the fourth part. In conclusion part, difference of PRG from two other methods and its benefits are explained.

## 2. AN OVERVIEW OF THE ROUGH SET THEORY

Rough Set Theory is based on an approach that is in order to define a set, unlike the classical set theory, in which set is defined by only its elements and no other information is given about the elements of set. In Rough Set Theory it is necessary to have some information about the elements of universe first. If objects are characterized with the same information, then they are same or indistinguishable. This relation of indistinguishability forms the base of Rough

Set Theory. The main problems that can be solved by Rough Set approach define the objects of the sets according to the property values, determining the dependence or partial dependence between properties, reducing properties, presenting the importance of properties and setting up the decision rules.,see Pawlak (1991). Moreover, Rough Set Theory can be used, for reducing data, discovering the dependencies, estimating the importance of data, setting up decision algorithms from data, classifying data, discovering patterns in data, finding similarity and difference between data and determining cause effect relations, see Pawlak and Slowinski (1994).

**2.1 Information System**

Data for Rough Set analysis is represented in a property-value table form in which each row shows an object or a sample and each column shows a property that qualifies an object. Property values belonging to objects are obtained by either measurement or human experiences. That kind of table is called Information System. An information system S is defined as *S = (U, A)*. *U* is non empty finite set of objects which is called *S*'s universal set. *A*, is non empty finite set of properties. Any $a \in A$ property is defined by $f_a : U \rightarrow V_a$ function. Set $V_a$ is called range set of *a*. The information systems that include decision information are called decision tables. Decision table is formed by adding decision information to existing information system. In this way, besides the properties of objects, the decisions belonging to these objects can be seen. In order to make this situation clearer, an example of information system and decision table can be examined. This example of decision table is formed by Komorowski and et. al (1998).

Table 2.1: An information system topic of which is people who applied jobs.

| **Person** | **Diploma** | **Experience** | **French** | **Reference** | **Decision** |
|---|---|---|---|---|---|
| $x_1$ | MBA | Medium | Yes | Excellent | Accept |
| $x_2$ | MBA | Low | Yes | Neutral | Reject |
| $x_3$ | MCE | Low | Yes | Good | Reject |
| $x_4$ | MSC | High | Yes | Neutral | Accept |
| $x_5$ | MSC | Medium | Yes | Neutral | Reject |
| $x_6$ | MSC | High | Yes | Excellent | Accept |
| $x_7$ | MBA | High | No | Good | Accept |
| $x_8$ | MCE | Low | No | Excellent | Reject |

Table 2.2: Numerical form of Table 2.1

| **Person** | **F1** | **F2** | **F3** | **F4** | **Decision** |
|---|---|---|---|---|---|
| $x_1$ | 1 | 2 | 1 | 3 | 1 |
| $x_2$ | 1 | 1 | 1 | 1 | 0 |
| $x_3$ | 2 | 1 | 1 | 2 | 0 |
| $x_4$ | 3 | 3 | 1 | 1 | 1 |
| $x_5$ | 3 | 2 | 1 | 1 | 0 |
| $x_6$ | 3 | 3 | 1 | 3 | 1 |
| $x_7$ | 1 | 3 | 2 | 2 | 1 |
| $x_8$ | 2 | 1 | 2 | 3 | 0 |

We can show the relation between U universe, *A* properties, *d* decision data and number values that belongs to objects as below.

$U = \{x_1, x_2, ..., x_8\}$
A = {F1, F2, F3, F4} = {Diploma, Experience, French, Reference}
d = Decision
Range set that belongs to properties:
F1 = {1; 2; 3}; 1 = MBA, 2 = MCE, 3 = MSC
F2 = {1, 2, 3}; 1 = Low, 2 = Medium, 3 = High
F3 = {1, 2}; 1= Yes, 2 = No
F4 = {1, 2, 3}; 1 = Neutral, 2 = Good, 3 = Excellent
d = {0, 1}; 0 = Reject, 1 = Accept

**2.2 Indiscernibility**

A decision table clarifies all information about information system. This table may be very large. Same or indiscernible objects may be shown more than one or some properties may be redundant.

If $S = (U, A)$ is an information system for any $B \subseteq A$, each subset of $B$ properties defines an equivalence relation in U universe. The name of this relation is indiscernibility relation. For different two objects in U universe, the equivalence relation $IND_S(B)$ defined in below is called B-indiscernibility relation.

$IND_S(B)=\{(x,y) \in U^2 \mid \forall a \in B \; a(x)=a(y)\}$

In indiscernebility relation $S$ index is omitted when it is clear that which information system is referred. If $(x,y) \in IND_{\mathbf{A}}(B)$, then $x$ and $y$ objects are indiscernible according to $B$. $x$ and $y$ objects are indiscernible because both of them have the same feature values and the decision can not be estimated. Before finding rule reducts in a decision table, it should be searched whether it has any indiscernible relations. Let Table 2.3 is handled after analyzing an information system. At first Table 2.3 is checked for indiscernible relations. It can be seen the objects $x_1, x_3$ and $x_4, x_5$ are indiscernible between each other. The reorganized decision table is shown in Table 2.4 below.

Table 2.3: A Sample Decision Table

| **Object** | **F1** | **F2** | **F3** | **F4** | **Decision** |
|---|---|---|---|---|---|
| $x_1$ | 1 | 2 | 1 | 3 | 1 |
| $x_2$ | 1 | 1 | 1 | 1 | 0 |
| $x_3$ | 1 | 2 | 1 | 3 | 1 |
| $x_4$ | 3 | 3 | 1 | 1 | 1 |
| $x_5$ | 3 | 3 | 1 | 1 | 1 |

Table 2.4: The Reorganized Decision Table that has no indiscernible relation

| **Object** | **F1** | **F2** | **F3** | **F4** | **Decision** |
|---|---|---|---|---|---|
| $x_1, x_3$ | 1 | 2 | 1 | 3 | 1 |
| $x_2$ | 1 | 1 | 1 | 1 | 0 |
| $x_4, x_5$ | 3 | 3 | 1 | 1 | 1 |

**2.3 Rule Reduct Generation**

One main concept in Rough Set Theory is rule reduct generation (RG). When there are a great variety of properties of objects, it is time consuming to control all properties. Especially when number of objects is high, solving the decision mechanism becomes very difficult. For example Table 2.2 shows the importance of rule reduction. When property F1 is equal to 2, then without looking the other properties, it is understood that the decision is equal to 0. Thus it can be said that if F1=2, then d=0. The case that F1=2, is rule reduct or r-reduct for the given information system. The other cases of the r-reduct are if F1=3 and F2=3, then d=1. Similarly, rule reduct can be looked for whole information system. There are many r-reducts in a decision table. When we consider that information systems in real life include much more data, importance of finding all rule reducts in short time can be understood clearly.

**2.4 Rule Reduct Generation(RG) Algorithm**

RG algorithm is proposed by Pawlak (1991) and modified by Kusiak and Tseng (1999). This algorithm tries to find all situations that consist of rule reduct. Steps of this algorithm are given below:

Step 0: Object number is defined $i=1$ and property number is defined $j=1$.

Step 1: For $k \neq i$, $j=1,..m$ is chosen. If $a_{ij} \neq a_{kj}$ and $a_{ij} = a_{kj} \wedge d_i = d_k$, then $a_{ij}$ is declared to be r-reduct. If it is tried for all properties of object, then step 2 is applied.

Step 2: $i=i+1$ is assigned. If it is tried for all objects, then step 3 is applied, if not step 1 is applied.

Step 3: Two properties are chosen and step 1 is applied. It works until m-1 property groups are tried and by this way all rule reducts are found.

**2.5 Modified Rule Generation (MRG) Algorithm**

Although RG algorithm detects all rule reducts, it cannot find minimal set of rule reducts. That is why redundant rule reducts may be found. Then work load increases and it causes to get the decision in a longer process. In r-reduct sample, in which the relation between the first property and decision is “2 x x x 0”, when $F_1 = 2$, d = 0. Nevertheless when we looked at two of the properties that hold “2 x x x 0” relation, it can be seen that if $F_1 = 2$ and $F_2 = 1$, d = 0, then the pair (F1,F2) is assigned to be r-reduct. Since F1 property is a r-reduct and it exists in the pair (F1,F2), it can be seen that (F1,F2) is redundant. Modified Rule Generation (MRG) algorithm is proposed by Guo and Chankong ( 2002). Aim of MRG algorithm is to find the minimal set of rule reducts. By this way, unnecessary operations are avoided and time needed to achieve result becomes shorter than RG algorithm. Steps of MRG can be summarized as below:

Step 0: Information system is sorted according to decision values.

Step 1: Object number is assigned as $i=1$ and property number in rule reduct is assigned as $r=1$.

Step 2: i$^{th}$ row is scanned from j=1. If $a_{ij} \neq "*"$ then step 3 is applied, if not then step 4 is applied.

Step 3: For all $k \neq i$, if $a_{ij} \neq a_{kj}$ or $a_{ij} = a_{kj} \wedge d_i = d_k$, then $a_{ij}$ is assigned to be r-reduct. If all columns are scanned for $j=1,..n$, then step 4 is applied, if not j is assigned to be $j=j+1$ and step 2 is applied.

Step 4: $i$ is assigned to be $i=i+1$ and step 2 is applied until the last object. When there is no object left, step 5 is applied.

Step 5: Decision table is revised according to objects which have same property value and properties involve $a_{ij} \neq "x"$ replace with "*" for 1-property reducts. Then step 6 is applied.

Step 6: In order to find higher degree rule reducts in revised $T'$ table, r is assigned to be $r = r+1$. If $r = m$, process is stopped, else $i$ is assigned to be $i = 1$ and step 7 is applied.

Step 7: By scanning i[th] row $a_{ij1},...a_{ijr}$ values, which belongs to $F_{j1},..., F_{jr}$ properties, it is controlled whether they fit r-property reduction or not. If a rule reduct is detected step is applied, if not step 8 is applied.

Step 8: Either for all $k \neq i$, if $j = j_1,...j_r$ or $a_{ij} \neq a_{kj}$ or for $a_{ij} = a_{kj}$, if $j = j_1,...j_r \wedge d_i = d_k$, then $\{a_{ij1},...a_{ijr}\}$ implies r-property rule reduct. $\{a_{ij1},...a_{ijr}\}$ property group, is indicated with "*r" in order to prevent from reuse. Step 7 is applied again.

Step 9: $i$ is assigned to be $i = i+1$. If $i$ is greater than the object number in U, then step 6 is applied, else step 7 is applied.

## 3. PRUNING RULE REDUCT GENERATION (PRG) ALGORITHM

PRG algorithm is developed within context of this paper in order to find a solution to the problem of finding rule reduct in information systems. It is modified to find minimal set of rule reduction cases faster than the other algorithms. This algorithm uses tree structured data type. Before comparing objects properties with others, it uses a tree of features to map the search. The tree is developed according to the features and it shows all possible subsets of the features. When a rule reduct is detected, it prunes the next related branch of tree in specific system. Thus, not only finding redundant rule reducts is avoided, but also by decrease in work load fewer comparisons are needed to make. It is an effective way of finding minimal set of rule reducts. Tree diagram of an algorithm used in four-property information system is as below (Figure 3.1):

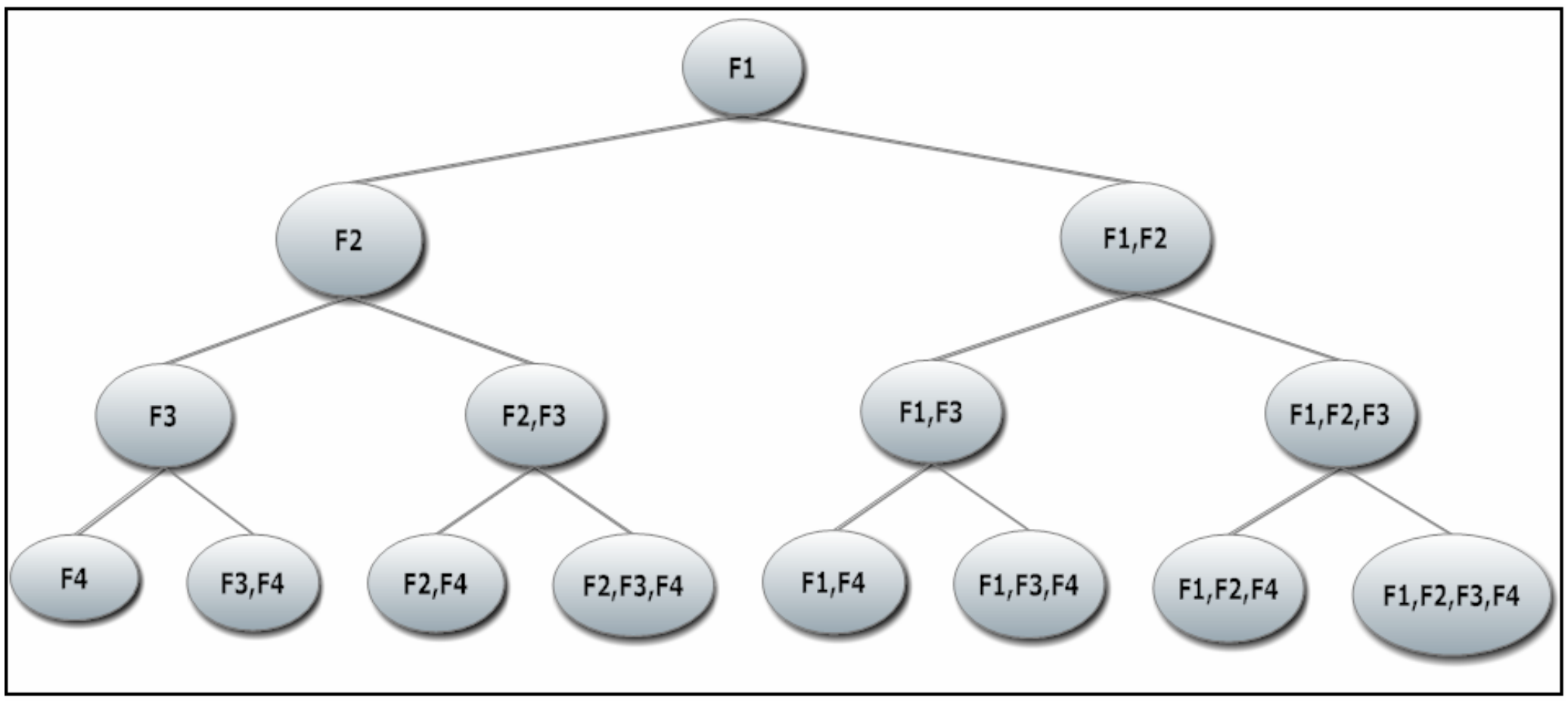

Figure 3.1: PRG algorithm for four-property tree data type

By looking Figure 3.1 it is seen that all subsets that belong to properties are located into tree. While examining rule reducts that belong to an object, in tree first of all, the way which will be followed starts from the root, and then goes to left child and finally the right child. Working principle of PRG algorithm can be told in detail as follows:

Step 0: Number of objects is assigned to be $i = 1$.

Step 1: By building up the tree, all keys in nodes are assigned to be k=0.

Step 2: Node = Root.

Step 3: SEQUENCEOFACTION (Node)

Step 4: Set $i = i + 1$. If all objects are done, go to step 5, if not go to step 1.

Step 5: Finish.

SEQUENCEOFACTION (Node)

IF (RULEREDUCTION (Node))

Node is assigned to be Node.key = 1

The Rule Reduct in node is declared.

For all nodes connected to right child of the node and all nodes that are in same line with the mentioned node and includes all properties that the node has, assign Node.key = 1. This process makes redundant branches of tree pruned.

If Node.left $\neq$ null

SEQUENCEOFACTION(Node.left)

If Node.right $\neq$ null and Node.key = 0

SEQUENCEOFACTION(Node.right)

RULEREDUCTION (Node)

If Node.key = 1 return "FALSE"

If for all j, $j \neq i$; [($a_{j,k_1} \neq a_{i,k_1}$) or ($a_{j,k_2} \neq a_{i,k_{21}}$) or $\ldots a_{j,k_t} \neq a_{i,k_t}$)] then the case $F_{k1} = a_{ik_1}, F_{k2} = a_{ik_2}, \ldots F_{kt} = a_{ik_t}, d = d_i$ is rule reduct. Return "TRUE"

If for all j, $j \neq i$; $[(a_{j,k_1} = a_{i,k_1})$ and $(a_{j,k_2} = a_{i,k_{21}})$ and… $a_{j,k_t} = a_{i,k_t})]$ and $d_j = d_i$ then the case $F_{k1} = a_{ik_1}, F_{k2} = a_{ik_2}, ... F_{kt} = a_{ik_t}, d = d_i$ is rule reduct. Return "TRUE"

Else return "FALSE"

## 4. COMPARISON PRG WITH RG AND MRG.

The Rule Reduct Generation (RG) and Modified Rule Generation (MRG) algorithms, that find rule reducts for an information system, are explained above. The Pruning Rule Generation (PRG) algorithm is developed as a more efficient and faster method to find minimal set of rule reducts. To compare the three methods, they are applied on Table 4.1 that is a decision table having four features, one decision value and five objects. This table and application of RG and MRG algorithms are taken from Guo and Chankong ( 2002)

Table 4.1: A Sample Decision Table

| **Object** | **F1** | **F2** | **F3** | **F4** | **Decision** |
|---|---|---|---|---|---|
| $x_1$ | 0 | 0 | 1 | 3 | 0 |
| $x_2$ | 0 | 1 | 1 | 1 | 1 |
| $x_3$ | 1 | 2 | 2 | 0 | 1 |
| $x_4$ | 0 | 1 | 0 | 2 | 2 |
| $x_5$ | 0 | 0 | 0 | 1 | 2 |

At first RG algorithm is executed to find rule reducts. As known RG algorithm finds all possible rule reducts and this causes a lot of redundant rule reducts. The result of RG algorithm is shown in Table 4.2. There are 50 possible one-feature, two-feature and three-feature rule reducts.

Table 4.2: All possible rule reducts (RG)

| | *F1* | *F2* | *F3* | *F4* | *d* | *obj* |
|---|---|---|---|---|---|---|
| One-feature r-reduct | NaN | NaN | NaN | 3 | 0 | 1 |
| | 1 | NaN | NaN | NaN | 1 | 3 |
| | NaN | 2 | NaN | NaN | 1 | 3 |
| | NaN | NaN | 2 | NaN | 1 | 3 |
| | NaN | NaN | NaN | 0 | 1 | 3 |
| | NaN | NaN | 0 | NaN | 2 | 4 |
| | NaN | NaN | NaN | 2 | 2 | 4 |
| | NaN | NaN | 0 | NaN | 2 | 5 |
| Two-feature r-reduct | 0 | NaN | NaN | 3 | 0 | 1 |
| | NaN | 0 | 1 | NaN | 0 | 1 |
| | NaN | 0 | NaN | 3 | 0 | 1 |
| | NaN | NaN | 1 | 3 | 0 | 1 |
| | NaN | 1 | 1 | NaN | 1 | 2 |
| | NaN | 1 | NaN | 1 | 1 | 2 |
| | NaN | NaN | 1 | 1 | 1 | 2 |
| | 1 | 2 | NaN | NaN | 1 | 3 |
| | 1 | NaN | 2 | NaN | 1 | 3 |
| | 1 | NaN | NaN | 0 | 1 | 3 |
| | NaN | 2 | 2 | NaN | 1 | 3 |
| | NaN | 2 | NaN | 0 | 1 | 3 |
| | NaN | NaN | 2 | 0 | 1 | 3 |
| | 0 | NaN | 0 | NaN | 2 | 4 |
| | 0 | NaN | NaN | 2 | 2 | 4 |
| | NaN | 1 | 0 | NaN | 2 | 4 |
| | NaN | 1 | NaN | 2 | 2 | 4 |
| | NaN | NaN | 0 | 2 | 2 | 4 |
| | 0 | NaN | 0 | NaN | 2 | 5 |
| | NaN | 0 | 0 | NaN | 2 | 5 |
| | NaN | 0 | NaN | 1 | 2 | 5 |
| | NaN | NaN | 0 | 1 | 2 | 5 |
| Three feature r-reduct | 0 | 0 | 1 | NaN | 0 | 1 |
| | 0 | 0 | NaN | 3 | 0 | 1 |
| | 0 | NaN | 1 | 3 | 0 | 1 |
| | NaN | 0 | 1 | 3 | 0 | 1 |
| | 0 | 1 | 1 | NaN | 1 | 2 |
| | 0 | 1 | NaN | 1 | 1 | 2 |
| | 0 | NaN | 1 | 1 | 1 | 2 |
| | NaN | 1 | 1 | 1 | 1 | 2 |
| | 1 | 2 | 2 | NaN | 1 | 3 |
| | 1 | 2 | NaN | 0 | 1 | 3 |
| | 1 | NaN | 2 | 0 | 1 | 3 |
| | NaN | 2 | 2 | 0 | 1 | 3 |
| | 0 | 1 | 0 | NaN | 2 | 4 |
| | 0 | 1 | NaN | 2 | 2 | 4 |
| | 0 | NaN | 0 | 2 | 2 | 4 |
| | NaN | 1 | 0 | 2 | 2 | 4 |
| | 0 | 0 | 0 | NaN | 2 | 5 |
| | 0 | 0 | NaN | 1 | 2 | 5 |
| | 0 | NaN | 0 | 1 | 2 | 5 |
| | NaN | 0 | 0 | 1 | 2 | 5 |

NaN denotes “x”.

Unlike to RG, MRG algorithm can eliminate the redundant rule reducts. It is remembered that MRG begins to find one-feature rule reducts. In Table 4.3 all one-feature rule reducts are shown. Then MRG makes a revision on decision table and replace one-feature rule reducts with '*'. The Table 4.4 is handled as a revised decision table of Table 4.3.

Table 4.3: One-feature rule reducts (MRG)

| F1 | F2 | F3 | F4 | d | obj |
|---|---|---|---|---|---|
| NaN | NaN | NaN | 3 | 0 | 1 |
| 1 | NaN | NaN | NaN | 1 | 3 |
| NaN | 2 | NaN | NaN | 1 | 3 |
| NaN | NaN | 2 | NaN | 1 | 3 |
| NaN | NaN | NaN | 0 | 1 | 3 |
| NaN | NaN | 0 | NaN | 2 | 4 |
| NaN | NaN | NaN | 2 | 2 | 4 |

Table 4.4: Revised decision table

| Obj.no | F1 | F2 | F3 | F4 | d |
|---|---|---|---|---|---|
| 1 | 0 | 0 | 1 | * | 0 |
| 2 | 0 | 1 | 1 | 1 | 1 |
| 3 | * | * | * | * | 1 |
| 4 | 0 | 1 | * | * | 2 |
| 5 | 0 | 0 | * | 1 | 2 |

After finding one-feature rule reducts, the decision table is revised then the revised table is used in MRG algorithm. Searching for two-feature rule reducts is executed and the two-feature rule reducts are shown in Table 4.5. As mentioned before, the decision table is revised again then Table 4.6 is reached.

Table 4.5: Two-feature rule reducts (MRG)

| F1 | F2 | F3 | F4 | d | obj |
|---|---|---|---|---|---|
| NaN | 0 | 1 | NaN | 0 | 1 |
| NaN | 1 | 1 | NaN | 1 | 2 |
| NaN | 1 | NaN | 1 | 1 | 2 |
| NaN | NaN | 1 | 1 | 1 | 2 |
| NaN | 0 | NaN | 1 | 2 | 5 |

Table 4.6: Revised decision table

| Obj.no | F1 | F2 | F3 | F4 | d |
|---|---|---|---|---|---|
| 1 | 0 | 0(*2) | 1(*2) | * | 0 |
| 2 | 0 | 1(*2,*2') | 1( *2,*2'') | 1(*2',*2'') | 1 |
| 3 | * | * | * | * | 1 |
| 4 | 0 | 1 | * | * | 2 |
| 5 | 0 | 0(*2) | * | 1(*2) | 2 |

The result of MRG algorithm is shown in Table 4.7. There are 12 possible one-feature and two-feature rule reducts. There is no three-feature rule reduct because all possible three-feature rule reducts are redundant. MRG algorithm eliminates all redundant rule reducts then it attains the minimal set of rule reducts as shown in Table 4.7.

Table 4.7: Minimal set of r-reducts (MRG)

| | *F1* | *F2* | *F3* | *F4* | *d* | *obj* |
|---|---|---|---|---|---|---|
| One feature r-reduct | NaN | NaN | NaN | 3 | 0 | 1 |
| | 1 | NaN | NaN | NaN | 1 | 3 |
| | NaN | 2 | NaN | NaN | 1 | 3 |
| | NaN | NaN | 2 | NaN | 1 | 3 |
| | NaN | NaN | NaN | 0 | 1 | 3 |
| | NaN | NaN | 0 | NaN | 2 | 4 |
| | NaN | NaN | NaN | 2 | 2 | 4 |
| Two feature r-reduct | NaN | 0 | 1 | NaN | 0 | 1 |
| | NaN | 1 | 1 | NaN | 1 | 2 |
| | NaN | 1 | NaN | 1 | 1 | 2 |
| | NaN | NaN | 1 | 1 | 1 | 2 |
| | NaN | 0 | NaN | 1 | 2 | 5 |

Pruning Rule Reduction (PRG) algorithm also eliminates redundant rule reducts. It finds minimal rule reducts for an information system. The difference between MRG and PRG is the elimination method. MRG needs to revise the decision table but PRG uses a tree structured data type. This tree acts as a map for searching rule reducts. While processing the method, some branches of the tree is pruned, hence it means there is a redundant rule reduct. The branches that are declared as redundant, are not searched for rule reduction. Therefore, processing time is shortened. Finally, minimal set of rule reduction is reached by PRG in a less time than MRG.

The fourth object is marked in Table 4.8. The PRG algorithm is tried to find the rule reducts for that object. The searching methodology is indicated step by step.

Table 4.8: Searching the fourth object for rule reduction

| **Object** | **F1** | **F2** | **F3** | **F4** | **Decision** |
|---|---|---|---|---|---|
| $x_1$ | 0 | 0 | 1 | 3 | 0 |
| $x_2$ | 0 | 1 | 1 | 1 | 1 |
| $x_3$ | 1 | 2 | 2 | 0 | 1 |
| $x_4$ | 0 | 1 | 0 | 2 | 2 |
| $x_5$ | 0 | 0 | 0 | 1 | 2 |

As it can be remembered PRG uses a tree structured data type. This tree that includes all subsets of the features is like a map for searching. In Figure 4.1 PRG tries to find a one-feature rule reduct for $x_4$. The *F1* value is 0 for $x_4$ and PRG decides there is no rule reduct for *F1* feature.

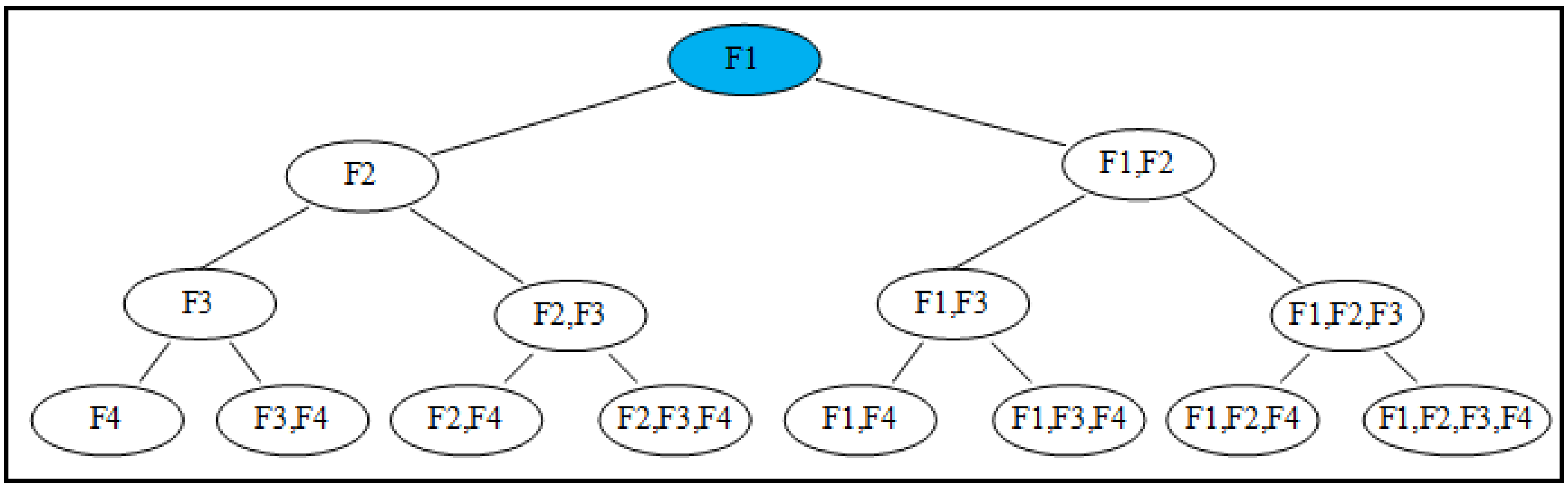

Figure 4.1: Searching one-feature rule reduct for $x_4$

Searching the rule reducts traces an in-order path on the tree. The second step is done in Figure 4.2. The algorithm decides that *F2* is not a rule reduct for $x_4$.

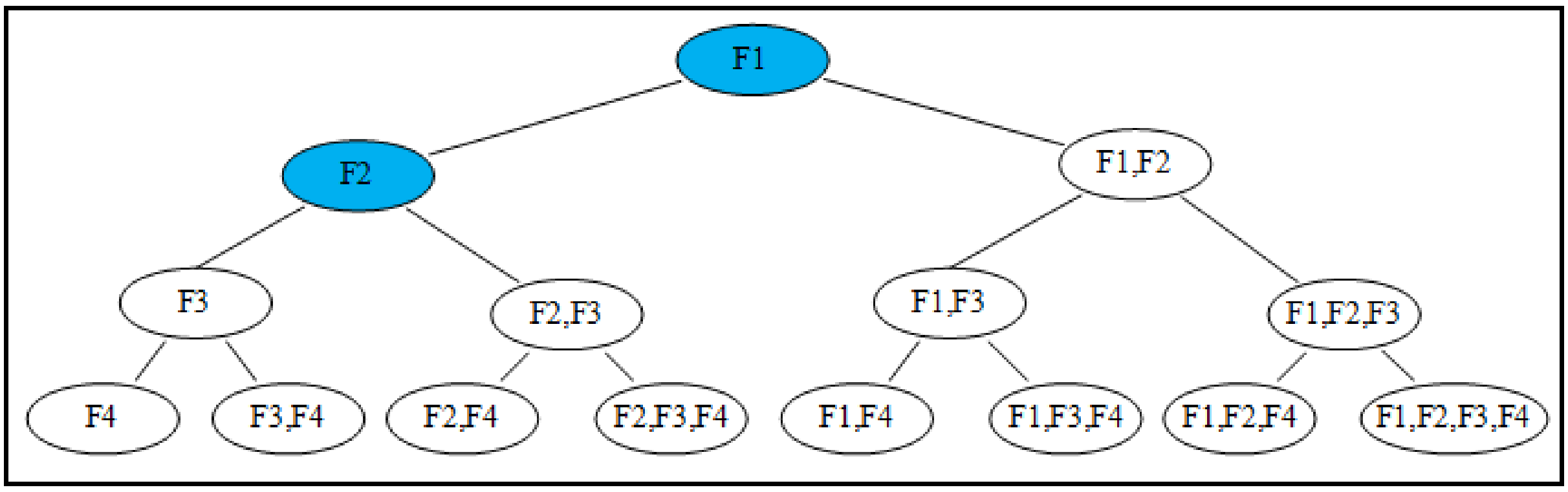

Figure 4.2: Searching one-feature rule reduct for $x_4$

The third step is done for *F3*. The value of *F3* is *0* and PRG algorithm determines that *F3* is a one-feature rule reduct for $x_4$. After finding the rule reduction, the tree branches are pruned. Pruning process is useful to avoid from redundant rule reducts. When a rule reduct is detected, the tree node that shows the feature group is marked. The right child of the node and all brother nodes that include the feature group are pruned. This pruning process provides less searching processes, shorter time to terminate and minimal rule reducts. In Figure 4.3 the tree node *F3* is declared as a one-feature rule reduct. After finding the rule reduction the tree nodes *F3F4*, *F2F3*, *F1F3*, *F1F2F3* are pruned, because they include *F3* feature and cause redundant rule reducts.

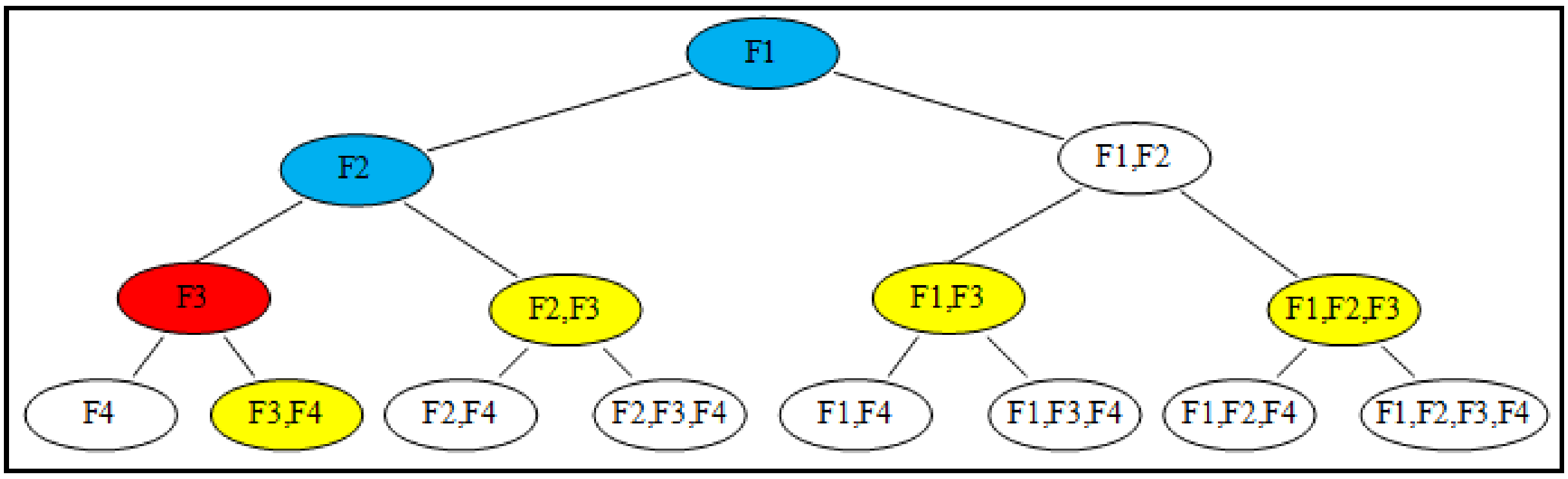

Figure 4.3: Finding one-feature rule reduct for $x_4$ and pruning next related tree branches

The next step is done for *F4* feature. PRG algorithm determines that F4 is a one-feature rule reduct for $x_4$. Then pruning process is begun and the tree nodes *F2F4*, *F2F3F4*, *F1F4*, *F1F3F4*, *F1F2F4*, and *F1F2F3F4* are pruned. Because these features include *F4* feature and they are redundant rule reducts.

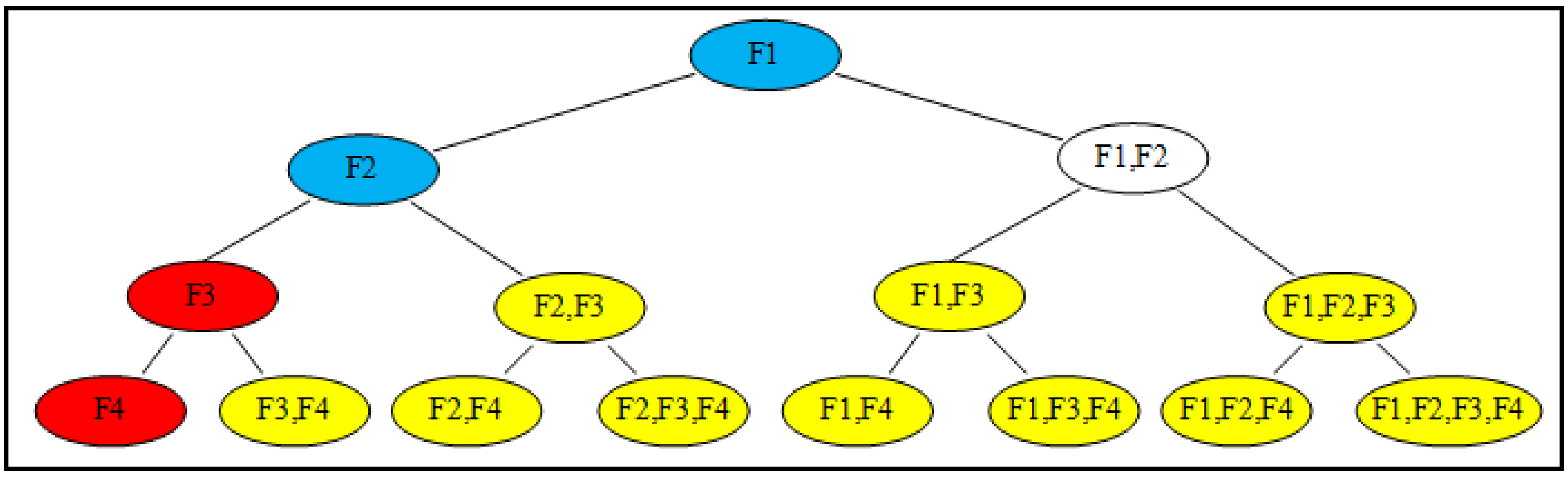

Figure 4.4: Finding one-feature rule reduct for $x_4$ and pruning next related tree branches

It is told above that the PRG traces the tree in-order direction. After finding rule reduct for *F4* feature, it is expected that the algorithm will begin to search for *F3F4* node. But since *F3* is a one-feature rule reduct, the algorithm passes the right child. Then the right child of the *F2*, *F2F3* will be searched, but it is also pruned so this node and its right child *F2F3F4* are passed. The next node is *F2F4*, but by reason of one-feature rule reduct for *F4*, it has been also pruned and it is passed. Finally in execution queue the node *F1F2* is searched whether it is a rule reduct or not. It is not rule reduct, hence the algorithm does not make any process for this feature couple. The result of execution is shown in Figure 4.5. The other six nodes are not searched by PRG because they have been all pruned.

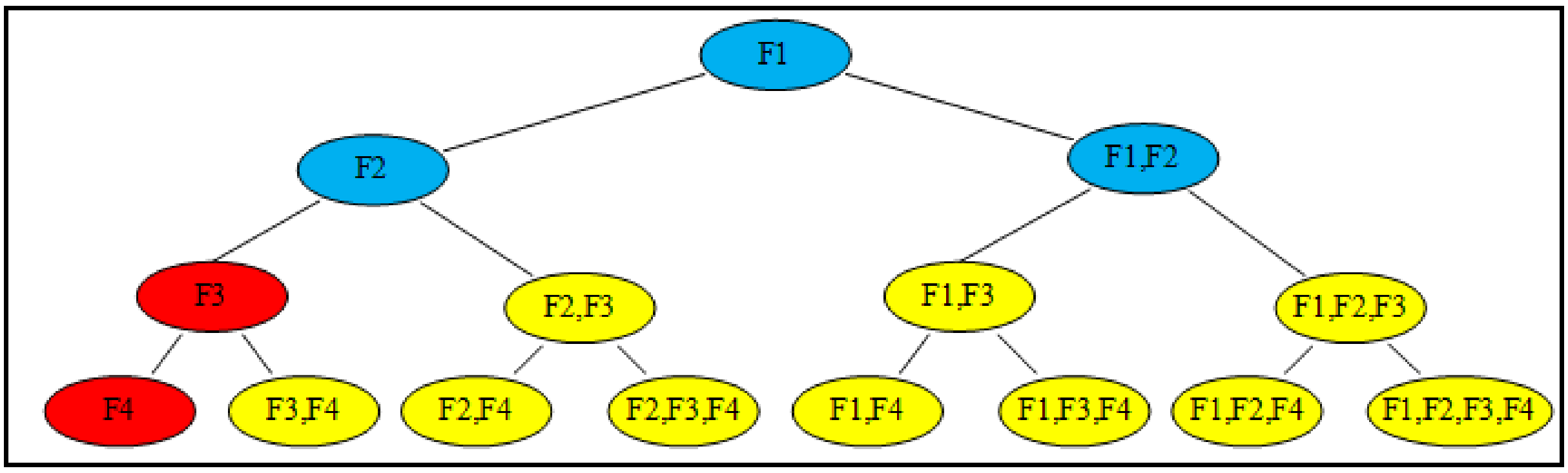

Figure 4.5: Searching two-feature rule reduct *F1F2* for $x_4$

## 5. CONCLUSION

Rough Set Theory presents quite successful solutions for analysis and classifications of big data sets that consist of a large number of qualifications. By rule reduction technique; analyzing data becomes easier. However, detecting rule reduction cases in information system of big data sets is pretty complicated. In this paper, information about methods of rule reduction called RG and MRG is given. Beside their availability, the cases, in which these two methods are deficient, are explained. Methods for problems are as effective as their availability. And, approaches with huge costs are rarely preferred. PRG algorithm, which is developed in this paper for more efficient and useable approaches, fills the deficient part of the other two algorithms. By avoiding redundant rule reductions, PRG algorithm not only facilitates the analysis of information system, but also makes the problem solved in shorter time.